\documentclass[sigconf]{acmart}

\usepackage{booktabs} 
\usepackage{microtype}
\usepackage{cleveref}
\usepackage{listings}
\usepackage{subcaption}
\usepackage{forest}
\usepackage{hyperref}

\usepackage{tikz}
\usetikzlibrary{arrows,er,positioning}
\usepackage{subcaption}
\usepackage{threeparttable}

\setcopyright{rightsretained}

\copyrightyear{2017} 
\acmYear{2017} 
\setcopyright{acmcopyright}
\acmConference{MM '17}{October 23--27, 2017}{Mountain View, CA, USA}\acmPrice{15.00}\acmDOI{10.1145/3123266.3129395}
\acmISBN{978-1-4503-4906-2/17/10}

\begin{document}
\setlength\abovecaptionskip{3truemm}

\title{ChainerCV: a Library for Deep Learning in Computer Vision}

\author{Yusuke Niitani}
\affiliation{%
  \institution{The University of Tokyo}
}
\email{niitani@jsk.imi.i.u-tokyo.ac.jp}

\author{Toru Ogawa}
\affiliation{%
  \institution{The University of Tokyo}
}
\email{t\_ogawa@hal.t.u-tokyo.ac.jp}

\author{Shunta Saito}
\affiliation{%
  \institution{Preferred Networks}
}
\email{shunta@preferred.jp}

\author{Masaki Saito}
\affiliation{
  \institution{Preferred Networks}
}
\email{msaito@preferred.jp}

\renewcommand{\shortauthors}{Y. Niitani, T. Ogawa, S. Saito, \& M. Saito}

\begin{abstract}
Despite significant progress of deep learning in the field of computer vision, there has not been a software library that covers these methods in a unifying manner.
We introduce ChainerCV, a software library that is intended to fill this gap.
ChainerCV supports numerous neural network models as well as software components needed to conduct research in computer vision.
These implementations emphasize simplicity, flexibility and good software engineering practices.
The library is designed to perform on par with the results reported in published papers and its tools can be used as a baseline for future research in computer vision.
Our implementation includes sophisticated models like Faster R-CNN and SSD, and covers tasks such as object detection and semantic segmentation.
\end{abstract}

%
%



\maketitle

\section{Introduction}
%

In recent years, the computer vision community has witnessed rapid progress thanks to deep learning methods in areas including image classification ~\cite{NIPS2012_4824}, object detection~\cite{Ren2015} and semantic segmentation~\cite{badrinarayanan2015segnet}.
High quality software tools are essential to keep up the rapid pace of innovation in deep learning research.
The quality of a software library is hugely influenced by traditional software quality metrics such as consistent coding conventions and coverage of tests and documentation.
In addition to that, a deep learning library, specifically a library that hosts implementations of deep learning models, needs to guarantee quality during the training phase.
Training a machine learning model to have a good performance is difficult due to numerous details that can hinder it from achieving its full potential.
This makes it all the more important that a library hosts implementations of high performance training code so that it can give guidance to developers and researchers who want to extend and develop further from these implementations.
We think that training code should perform on par with the performance reported by the paper that the implementation is based on.
On top of providing a high quality implementation for training a model, we also aim at making it more accessible, especially for users with limited experience, to run inference on sophisticated computer vision models such as Faster R-CNN~\cite{Ren2015}.

The rapid progress of deep learning research has been enabled by a number of frameworks including Chainer~\cite{Tokui2015}, TensorFlow~\cite{tensorflow2015-whitepaper} and PyTorch \footnote{\url{http://pytorch.org}}.
These frameworks have supported fundamental components of deep learning software such as  automatic differentiation and effective parallelization using GPUs.
However, they are intended to target general usage, and do not aim to provide complete implementations of vision algorithms.

Our software, ChainerCV, supports algorithms to solve tasks in the computer vision field such as object detection, while considering usability and predictable performance as the top priorities.
This makes it perfect to be used as a building block in larger software projects such as robotic software systems even by developers who are not computer vision experts.
Recently there has been a growing trend of building new neural network models using existing architectures as building blocks.
Examples can be seen in tasks such as instance segmentation~\cite{mask_rcnn} and scene graph generation~\cite{xu2017scenegraph}, which depend on object detection algorithms to localize objects in images. 
ChainerCV's algorithms can be used as components to construct software that can solve complex computer vision problems.

Training a network is a critical part of a machine learning algorithm, and ChainerCV is designed to make this process easy.
In many use cases, users need a machine learning model to perform well on a particular dataset they have.
Often a pretrained model is not sufficient for the users' tasks, and so they must re-train the model using their datasets.
To make training a model easier in such cases, ChainerCV provides reference implementations to train models, which can be used as a baseline to write new training code.
In addition to that, pretrained models can be used together with the users dataset to fine-tune the model.
ChainerCV also provides set of tools for training a model including dataset loader, prediction evaluator and visualization tools.

Reproducibility in machine learning and computer vision is one of the most important factors affecting the quality of the research.
ChainerCV aims at easing the process of reproducing the published results by providing training code that is guaranteed to perform on par with them.
These algorithms would serve as baselines to find a new idea through refinement and as a tool to compare a new approach against existing approaches.

To summarize, ChainerCV offers the following two contributions:
\begin{itemize}
  \item High quality implementations of deep learning-based computer vision algorithms to solve problems with emphasis on usability.
  \item Reference code and tools to train models, which is guaranteed to perform on par with the published results.
\end{itemize}
Our code is released at \url{https://github.com/pfnet/chainercv}.

\section{Related Work}

Deep learning frameworks such as Chainer~\cite{Tokui2015} and TensorFlow~\cite{tensorflow2015-whitepaper} play a fundamental role in deep learning software.
However, these software packages focus on fundamental components such as automatic differentiation and GPU support.
Keras~\cite{chollet2015keras} is a high-level deep learning API that is intended to enable fast experimentation.
While ChainerCV shares a similar goal with Keras to enable fast prototyping, our software provides more thorough coverage of software components for the computer vision tasks.
In addition to that, Keras does not provide high performance training code for sophisticated vision models like Faster R-CNN~\cite{Ren2015}.

OpenCV~\cite{itseez2015opencv} is a prominent example of computer vision software libraries supporting numerous highly tuned implementations.
The library supports wide range of algorithms including some deep learning-based algorithms, which emphasize on running inference on a cross platform environment.
Different from their work, ChainerCV aims at acceralating research in this field in a more comprehensive manner by
providing high quality training code on top of implementations to conduct inference.

Orthogonal to our open source work, there are several proprietary software solutions that support computer vision algorithms based on deep learning.
These include Computer Vision System Toolbox by MATLAB and Google Cloud Vision API by Google Cloud Platform.

Model Zoo hosts a number of open source implementations and their trained models.
Algorithms for a wide range of tasks are hosted on their website, but they are not provided as a library that organizes code in some standardized manner.
There are open source implementations released by the authors of papers and third-party implementations released by open source developers.
The primarily aim of these works is to make a prototype of a research idea.
Unlike them, one of our goals is to develop an implementation that follows a good software engineering practices so that
it is readable and easily extendable to other projects.
We assure the quality by developing through peer review process, and thorough coverage of documentations and tests.

More closely related to our work is \textbf{\it pytorch/vision}, which is a computer vision library that uses PyTorch as its backend.
Similar to our work, it hosts pretrained models to let users use high performance convolutional neural networks off-the-shelf.
At the time of writing, its support for pretrained model and data preparation are limited only to classification tasks.

\section{Implementation}
\begin{figure}[t]
\fbox{\centering\medskip
\begin{forest}
  for tree={
    font=\ttfamily,
    grow'=0,
    child anchor=west,
    parent anchor=south,
    anchor=west,
    calign=first,
    edge path={
      \noexpand\path [draw, \forestoption{edge}]
      (!u.south west) +(7.5pt,0) |- node[fill,inner sep=1.25pt] {} (.child anchor)\forestoption{edge label};
    },
    before typesetting nodes={
      if n=1
        {insert before={[,phantom]}}
        {}
    },
    fit=band,
    before computing xy={l=15pt},
  }
[.
  [chainercv
    [datasets:  Dataset loaders]
    [evaluations:  Evaluation metrics]
    [links:  Neural network models]
    [transforms:  Transform functions]
    [visualizations:  Visualization functions]
  ]
  [examples:  Training and demo codes for networks]
]
\end{forest}
}
\caption{Directory structure of ChainerCV}
\label{fig:directory}
\end{figure}
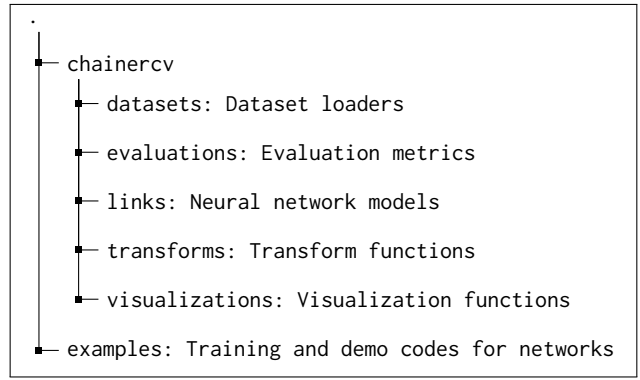
\subsection{Task Specific Models}
Currently, ChainerCV supports networks for object detection and semantic segmentation (\Cref{fig:directory}).
Object detection is the task of finding objects in an image and classifying them.
Semantic segmentation is the task of segmenting an image into pieces and assigning object labels to them.

We implemented our detection algorithms in a unifying manner by exploiting the fact that many of the leading state of the art architectures have converged on a similar structure~\cite{Huang2016}.
All of these architectures use convolutional neural networks to extract features and use sliding windows to predict localization and classification.
Our implementation includes architectures that can be grouped by Faster R-CNN~\cite{Ren2015} and Single Shot Multibox Detector (SSD)~\cite{Liu} meta-architectures.
Faster R-CNN takes a crop proposed by an external neural network called Region Proposal Networks and carry out classification on the crop of the input image.
SSD tries to alleviate the extra time running Region Proposal Networks by directly predicting classes and coordinates of bounding boxes. 
These meta-architectures are instantiated into more concrete networks that have different feature extractors or different head architectures.
These different implementations inherit from the base class for each meta-architecture using our flexible class design.

Our implementation of semantic segmentation models includes SegNet~\cite{badrinarayanan2015segnet}.
The architecture follows an encoder-decoder style.
We have separated a module to calculate loss from a network that predicts a probability map.
This design makes the loss reusable in other implementation of semantic segmentation models, which we are planning to add in the future.

Models for a certain task are designed to have a common interface.
For example, detection models support a {\tt predict} method that takes images and outputs bounding boxes around regions where objects are predicted to be located.
The common interface allows users to swap different models easily inside their code.
On top of that, the common interface is necessary to build functions that interact with neural network models by passing input images and receiving predictions.
For instance, thanks to this interface, we can write a function that iterates over a dataset and visualizes the predictions of all the samples.

\subsection{Datasets}
In order to train and evaluate deep learning models, datasets are needed.
ChainerCV provides an interface to datasets commonly used in computer vision tasks, such as datasets from the Pascal VOC Challenge~\cite{Everingham10}.
The datasets object downloads data from the Internet if necessary, and returns requested contents with an array-like interface.

\subsection{Transforms}
A transform is a function that takes an image and annotations as inputs and applies a modification to the inputs such as image resizing.
These functions are composed together to create a custom data preprocessing pipeline.

ChainerCV uses {\tt TransformDataset} to compose different transforms.
This is a class that wraps around a dataset by applying a function to a sample retrieved from the underlying dataset, which is often prepared to simply load data from a file system without any modifications.



We found that extending a dataset with an arbitrary function is effective especially in the case where multiple objects are processed in an interdependent manner.
Such interdependence of transforms happen in a scenario when an image is randomly flipped horizontally to augment a dataset and coordinates of bounding boxes are altered depending on whether the image is flipped or not. 
See \Cref{fig:transform} for the code to carry out the data preprocessing pipeline.
\begin{figure}
\begin{lstlisting}[language=Python]
from chainercv.datasets import *
from chainercv.transforms import *

dataset = VOCDetectionDataset()

def flip_transform(in_data):
    img, bbox, label = in_data
    img, param = random_flip(
        img, x_flip=True,
        return_param=True)
    bbox = flip_bbox(
        bbox, x_flip=param['x_flip'])
    return img, bbox, label

new_dataset = TransformDataset(
    dataset, flip_transform)
\end{lstlisting}
\caption{Code to preprocess data using {\tt TransformDataset}}
\label{fig:transform}
\end{figure}



\begin{figure*}[ht]
  \begin{minipage}{0.33\linewidth}
  	\centering
  	\includegraphics[width=\linewidth, trim=15 20 10 5, clip]{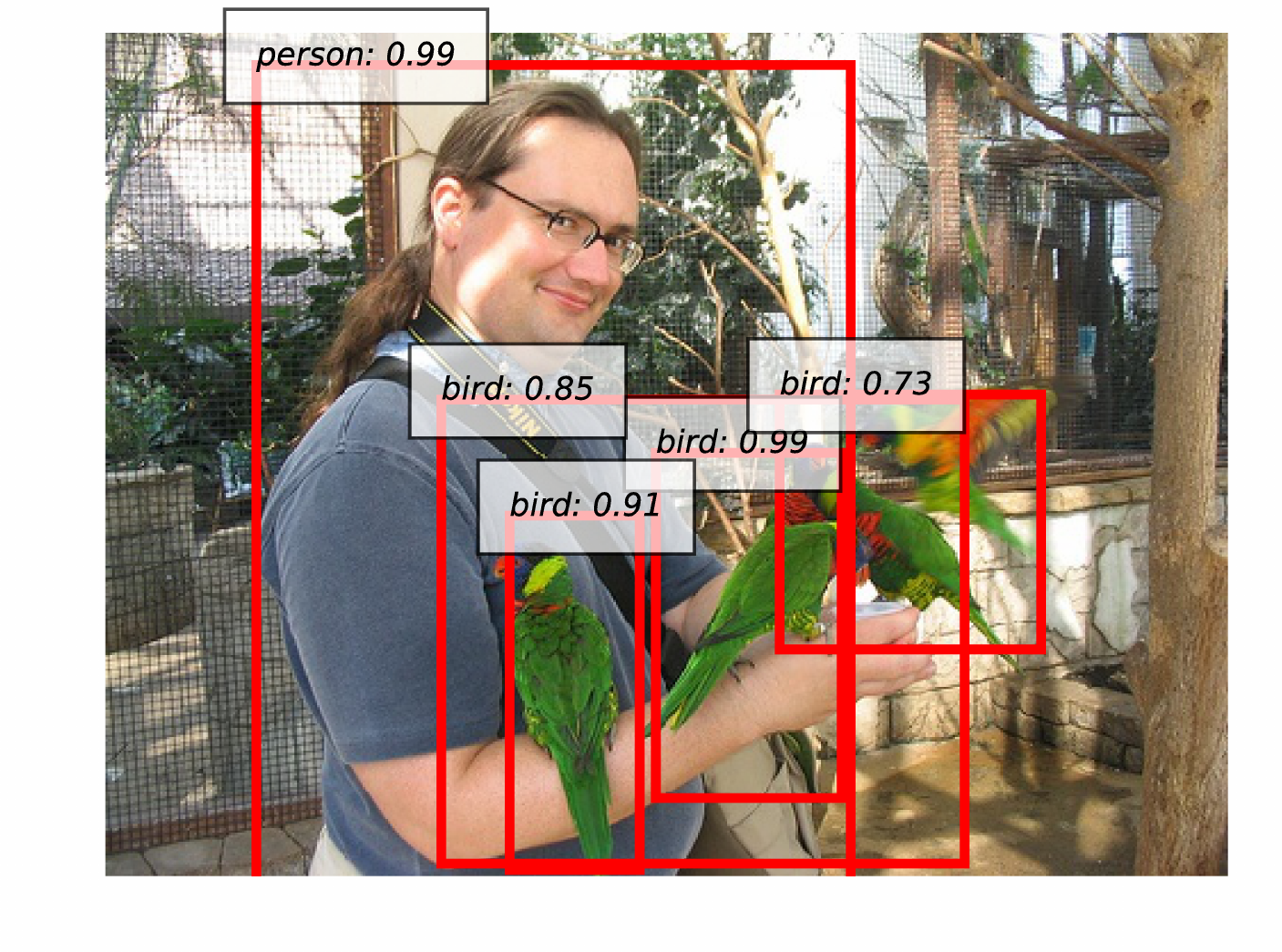}
    \includegraphics[width=\linewidth, trim=15 20 10 5, clip]{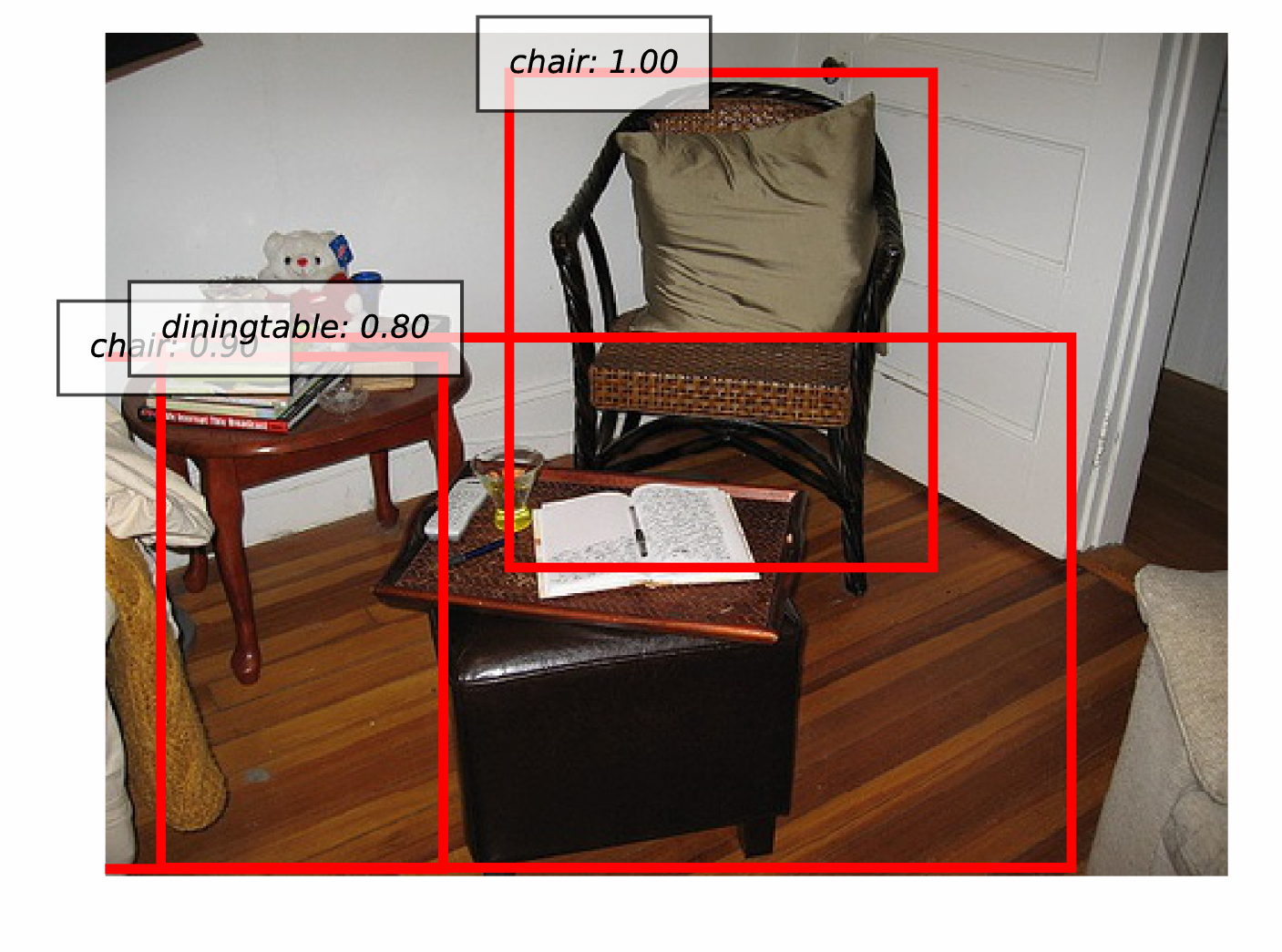}
  	\subcaption{Faster R-CNN}\label{fig:vis_faster_rcnn}
  \end{minipage}
  \begin{minipage}{0.33\linewidth}
    \centering
  	\includegraphics[width=\linewidth, trim=15 20 10 5, clip]{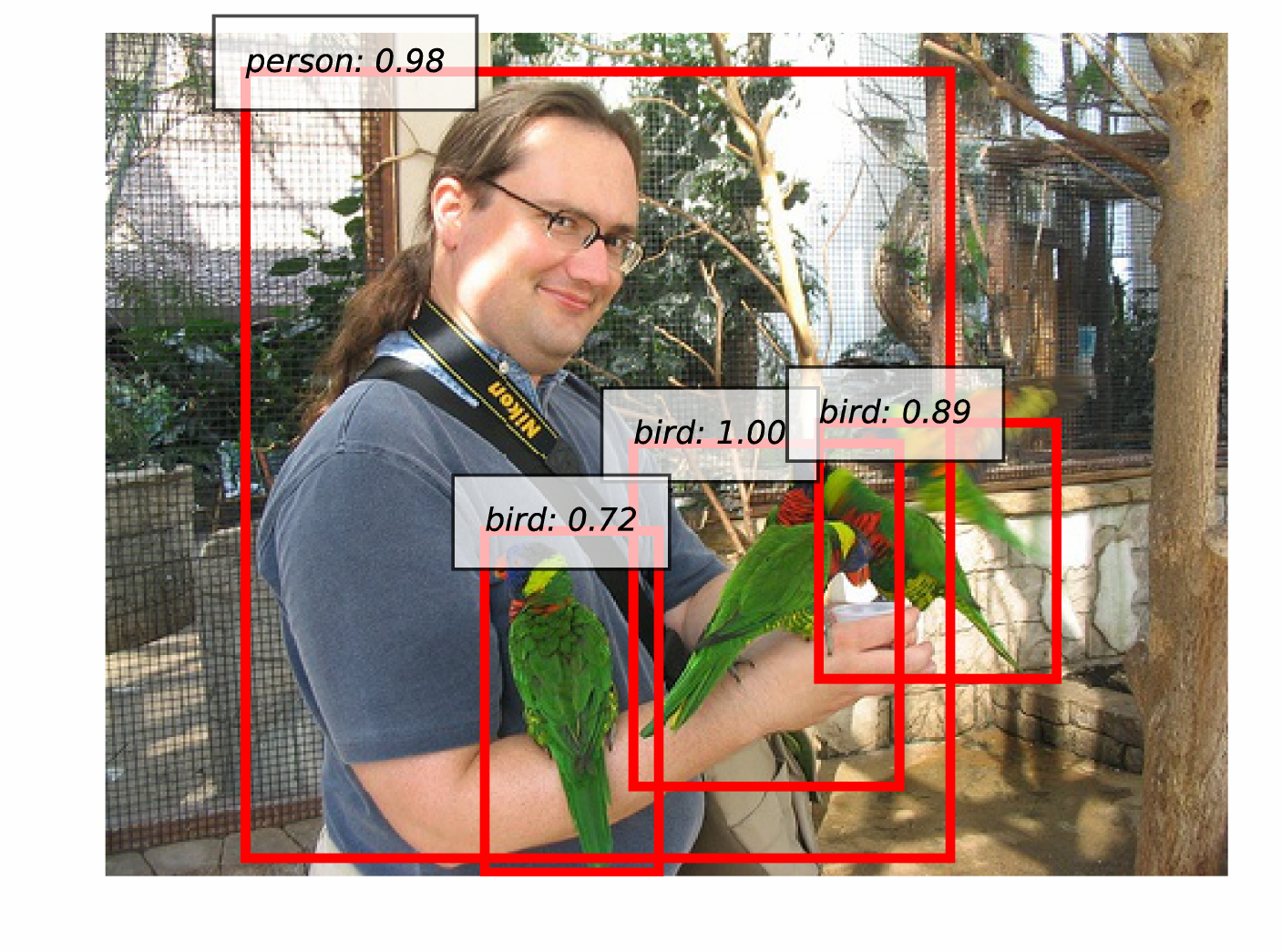}
   	\includegraphics[width=\linewidth, trim=15 20 10 5, clip]{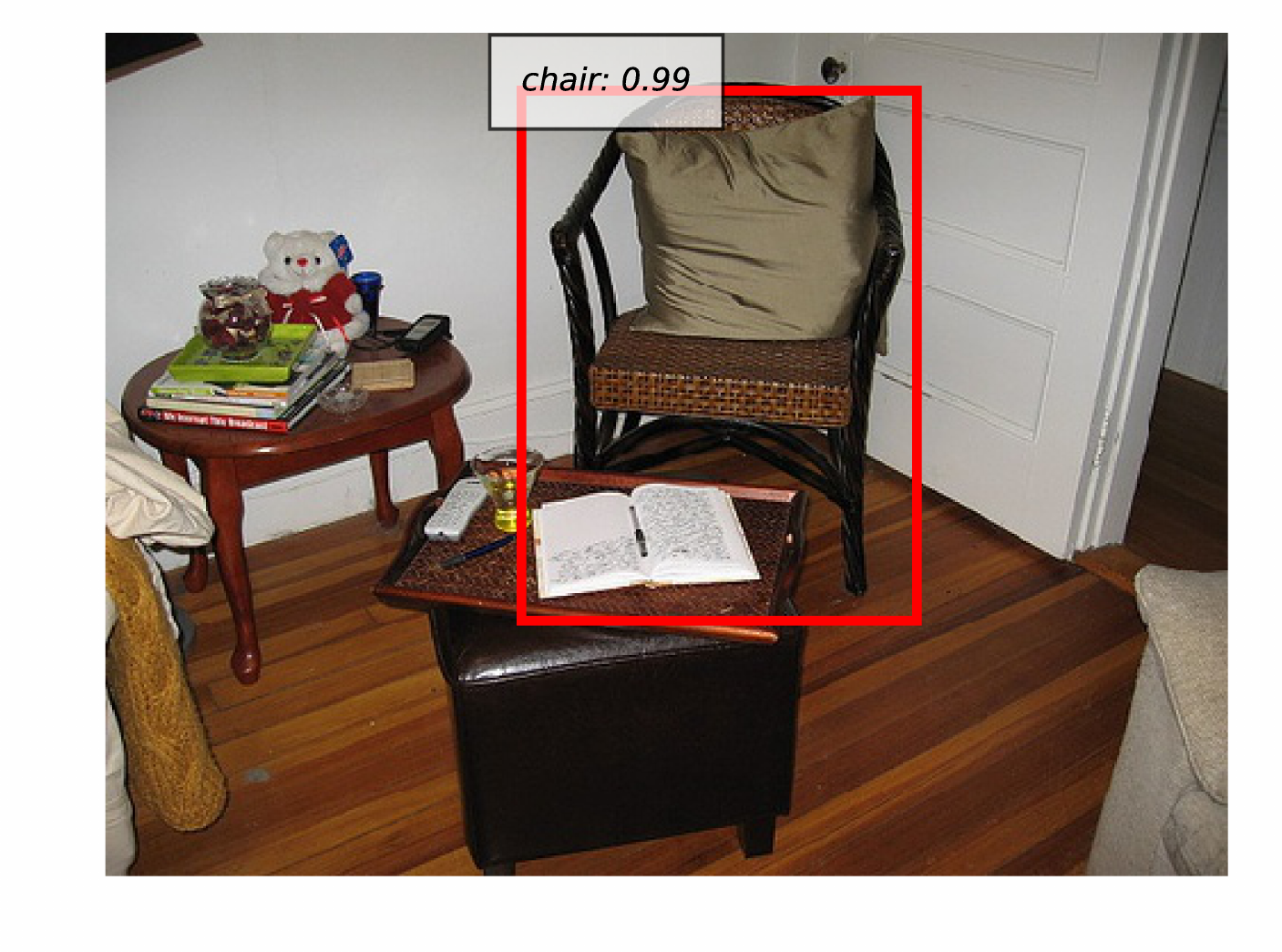}
	\subcaption{SSD512}\label{fig:vis_ssd}
  \end{minipage}
  \begin{minipage}{0.33\linewidth}
    \centering
  	\includegraphics[width=\linewidth, trim=15 20 5 5, clip]{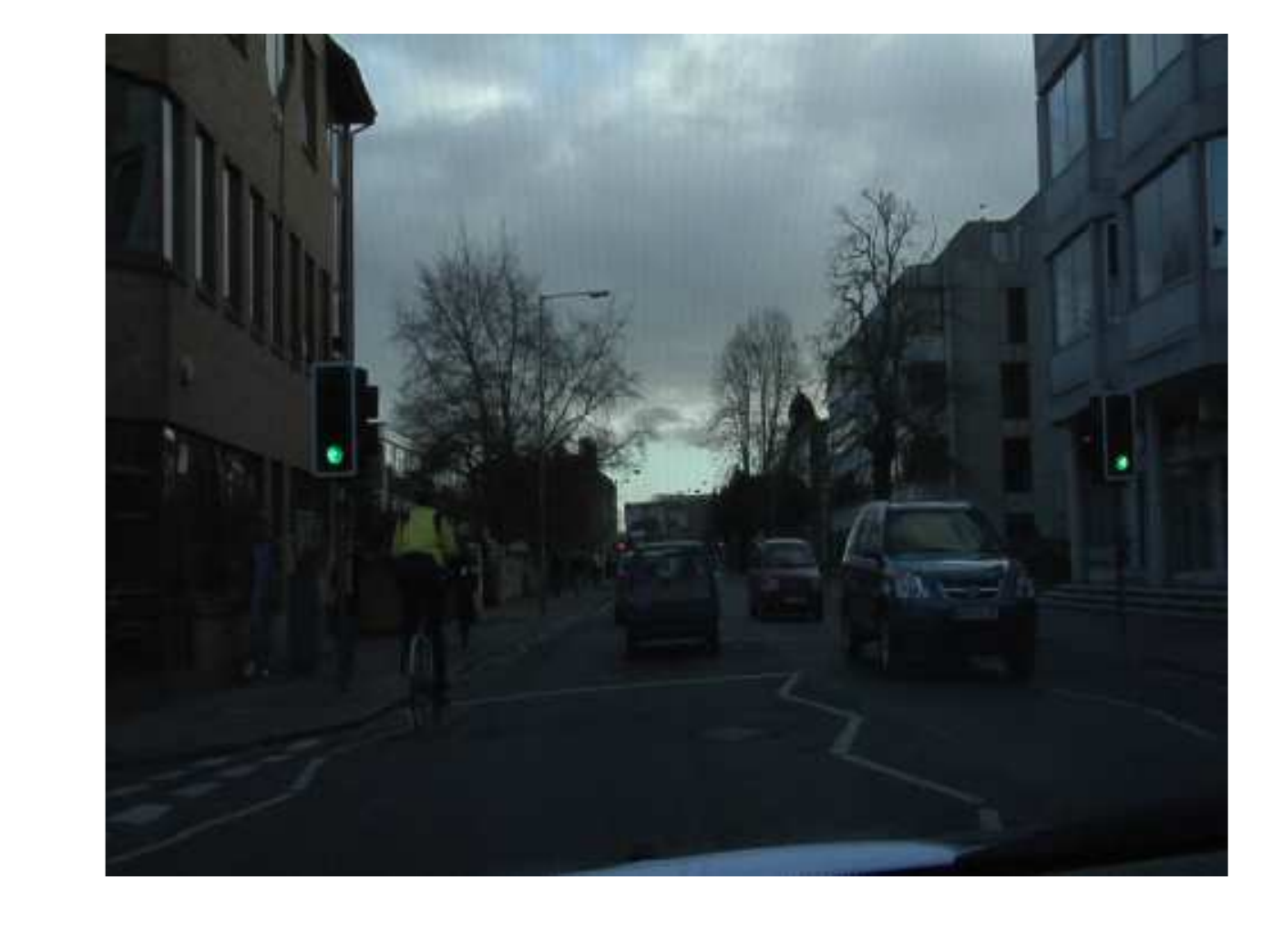}
    \includegraphics[width=\linewidth, trim=15 20 5 5, clip]{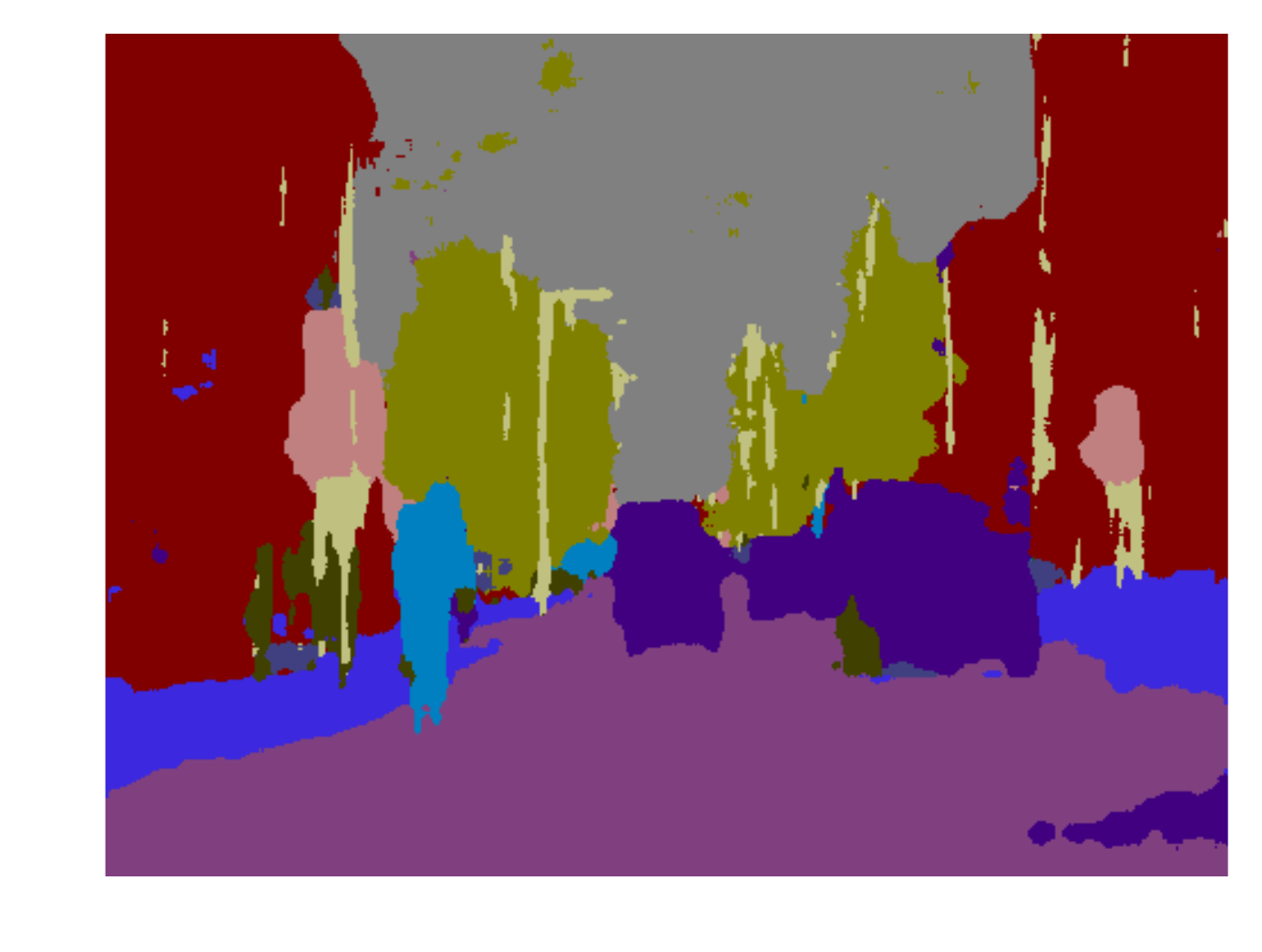}
  	\subcaption{SegNet (top: input, bottom: output)}\label{fig:vis_segnet}
  \end{minipage}
  \caption{Example visualizations. (\subref{fig:vis_faster_rcnn},  \subref{fig:vis_ssd}) Visualizations of object detection and (\subref{fig:vis_segnet}) visualizations of semantic segmentation.}
  \label{fig:vis_models}
\end{figure*}

\subsection{Visualizations and Evaluation Metrics}
ChainerCV supports a set of functions for visualization and evaluation, which are important for conducting research in computer vision.
These functions can be used across different models by enforcing a consistent data representation for each type of data.
For example, an image array is assumed to be RGB and shaped as \texttt{(C, H, W)}, where the elements of the tuple are channel size, height and width of the image.
The evaluations and visualizations in \Cref{sec:experiments} are carried out using the functions in ChainerCV.

%
%

\subsection{GPU Arrays}

As done in Chainer~\cite{Tokui2015}, ChainerCV uses {\tt cupy.ndarray} to represent arrays stored in GPU memory and {\tt numpy.ndarray} to represent arrays stored in CPU memory.
CuPy \footnote{\url{https://github.com/cupy/cupy}} is a NumPy like multi-dimensional array library with GPU acceleration.
Many functions in ChainerCV support both types as arguments, and returns the output with the same type as the input.
These functions include non-maximum suppression~\cite{7471831}, which is efficiently implemented using CUDA parallelization.
It is often complicated to set up a library to call CUDA kernels from a python module because the installation needs to consider a variety of machine configurations.
ChainerCV relies on CuPy when calling CUDA kernels, and its installation procedure is quite simple.

\section{Experiments}
\label{sec:experiments}
We report performance of the implemented training code, and verify that the scores are on par with the ones reported in the original papers.
Note that due to randomness in training, it is inevitable to produce slightly different scores from the original papers.

\subsection{Faster R-CNN}
We evaluated the performance of our implementation of Faster R-CNN, and compared it to the performance reported in the original paper~\cite{Ren2015}.
We experimented with a model that uses VGG-16 model~\cite{Simonyan14c} as a feature extractor.
The model is trained on the PASCAL VOC detection dataset.
The model is trained on the 2007 \texttt{trainval} and evaluated on 2007 \texttt{test} using our training code.
Some detection results of this trained model are shown in \Cref{fig:vis_faster_rcnn}.
The performance is compared against the original implementation using mean average precision in \Cref{table:detection}. 
Due to the stochastic training process, it is known that the final performance fluctuates~\cite{chen17implementation}.


\subsection{Single Shot Multibox Detector (SSD)}
We evaluated the performance of our implementations of SSD300 and SSD512, and compared them to the performance reported in~\cite{Fu2017}.
We trained these models with the trainval splits of PASCAL VOC 2007 and 2012 for training.
The performance is compared against the original implementation using mean average precision in \Cref{table:detection}. Note that we changed the train batchsize of SSD512 from 32 to 24 due to the GPU
memory limitation.
We also show some detection results of SSD512 in \Cref{fig:vis_ssd}.
\begin{table}[t]
  \caption{Object detection mean average precision (\%).}
  \label{table:detection}
  \begin{tabular}{cccl}
	\toprule
    Implementations & Faster R-CNN & SSD300 & SSD512 \\
    \midrule
    Original & 69.9 \cite{Ren2015} & 77.5 \cite{Fu2017} & 79.5 \cite{Fu2017} \\
    ChainerCV & 70.5 & 77.5 & 80.1 \\
    \bottomrule
  \end{tabular}
\end{table}

\subsection{SegNet}
We evaluated the performance of our SegNet implementation, and compared it to the performance reported in the journal version of the original paper~\cite{badrinarayanan2015segnet}. It is trained on the train split of CamVid~\cite{badrinarayanan2015segnet}, and evaluated on the test split.
The performance is measured by pixel accuracy, mean pixel accuracy and mean IoU, which are the metrics used in ~\cite{badrinarayanan2015segnet}.
The score is shown in~\Cref{table:semantic} and an example result is shown in~\Cref{fig:vis_segnet}.
\begin{table}[t]
\caption{Semantic segmentation performance of SegNet.}
  \label{table:semantic}
  \begin{tabular}{cccl}
    \toprule
    & pixel accuracy &mean pixel accuracy& mIoU \\
    \midrule
    Original \cite{badrinarayanan2015segnet} & 82.7 & 62.3 & 46.3 \\
    ChainerCV & 82.8 & 67.1 & 47.2 \\
  \bottomrule
\end{tabular}
\end{table}

\section{Conclusion}
In this article we have introduced a new computer vision software library that focuses on deep learning-based methods.
Our software lowers the barrier of entry to use deep learning-based computer vision algorithms by providing a convenient and unified interface.
It also provides evaluation and visualization tools to aid research and development in the field.
Our implementation achieves performance on par with the reported results, and we expect it to be used as a baseline to be extended with new ideas.

\section*{Acknowledgement}
We would like to thank Richard Calland for helpful discussion.

\bibliographystyle{ACM-Reference-Format}
\bibliography{sigproc} 

\end{document}